\documentclass{article}
\usepackage{iclr2027_conference,times}

\usepackage{amsmath,amsfonts,bm}

\def\1{\bm{1}}

\DeclareMathAlphabet{\mathsfit}{\encodingdefault}{\sfdefault}{m}{sl}
\SetMathAlphabet{\mathsfit}{bold}{\encodingdefault}{\sfdefault}{bx}{n}

\usepackage{amsmath}
\usepackage{amssymb}
\usepackage{marvosym}
\usepackage{booktabs}
\usepackage{flafter}
\usepackage{graphicx}
\usepackage[colorlinks=true,allcolors=blue]{hyperref}
\usepackage{multirow}
\usepackage{placeins}
\usepackage{url}
\usepackage{wrapfig}
\usepackage{xcolor}
\usepackage{colortbl}

\definecolor{scoregraylight}{RGB}{230,230,230}
\definecolor{scoregraydark}{RGB}{205,205,205}
\definecolor{scoregreen}{RGB}{0,166,81}

\hypersetup{hidelinks}
\title{ConDenseDrive: Knowledge-Dense, Token-Sparse Visual Encoder for End-to-End Autonomous Driving}
\author{Shengzhen Li\textsuperscript{1,2}, Yongkang Li\textsuperscript{1}, Hao Gao\textsuperscript{1}, Chenxi Wu\textsuperscript{1,2} \\
\textbf{Yajing Sun\textsuperscript{2}, Lu Hou\textsuperscript{2}, Wenyu Liu\textsuperscript{1}, Xinggang Wang\textsuperscript{1,\Letter}} \\
\normalfont
\textsuperscript{1}Huazhong University of Science and Technology \\
\textsuperscript{2}Yinwang Intelligent Technology Co. Ltd. \\
\texttt{\{shengzhenli,xgwang\}@hust.edu.cn}
}

\iclrfinalcopy

\begin{document}

\maketitle
\fancyhead{}
\raggedbottom

\begin{abstract}
End-to-end driving requires rich spatiotemporal representations to support scene understanding and trajectory planning under strict latency constraints. However, most systems initialize their visual encoders from generic image pretraining and densely process all patch tokens for each camera view and input frame, limiting both representation quality and inference efficiency. To address these challenges, we present ConDenseDrive, a knowledge-dense and token-sparse visual encoder for end-to-end autonomous driving. Our three-stage framework first makes the encoder knowledge-dense by distilling complementary visual, semantic, and geometric knowledge from three vision foundation models. It then adapts the distilled encoder to multi-frame end-to-end driving and integrates spatiotemporal context into register tokens for planning. Finally, it makes the encoder token-sparse by applying register-guided token compression, reducing redundant patch computation and yielding a \(2.0\times\) speedup while improving planning performance. ConDenseDrive achieves \(94.5\) PDMS on NAVSIM v1 and \(90.8\) EPDMS on NAVSIM v2, and reaches \(37.4\) HD-Score in zero-shot evaluation on HUGSIM. Extensive experiments and qualitative analyses demonstrate that visual encoder design provides an effective route to reliable and efficient end-to-end driving.

\end{abstract}

\section{Introduction}
\label{sec:introduction}

\begin{figure}[t!]
    \centering
    \includegraphics[width=0.9\linewidth]{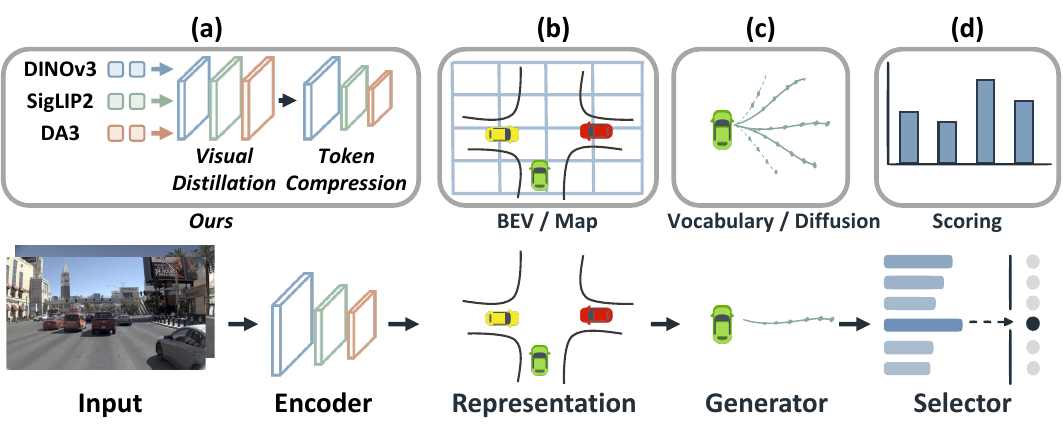}
\caption{\textbf{Design focuses in end-to-end driving.}
    The bottom row outlines an end-to-end driving pipeline.
    Representative methods focus on (b) scene representation~\citep{hu2023uniad,jiang2023vad}, (c) trajectory generation~\citep{jiang2026vadv2,liao2025diffusiondrive}, and (d) trajectory selection~\citep{yao2025drivesuprim,li2025gtrs}.
    ConDenseDrive instead targets (a) visual encoding through multi-teacher distillation and token compression.}
    \label{fig:paradigm}
\end{figure}

End-to-end autonomous driving predicts future ego trajectories from sensor observations and navigation conditions~\citep{chitta2023transfuser,hu2023uniad}. Recent end-to-end autonomous driving systems have advanced structured scene representations~\citep{hu2023uniad,jiang2023vad}, trajectory generation~\citep{jiang2026vadv2,liao2025diffusiondrive}, and trajectory scoring and selection~\citep{yao2025drivesuprim,li2025gtrs,gao2026rad2}, as illustrated in Fig.~\ref{fig:paradigm}. These methods rely on visual encoders, which supply scene information to the planner and can contribute substantially to inference latency (Fig.~\ref{fig:efficiency-tradeoff}(b)). Despite this central role, visual encoders are typically initialized from generic image pretraining. Their representation quality and computational cost are rarely considered jointly.

A straightforward solution is to scale the visual backbone to accommodate richer representations, but the added computation increases visual encoding latency. Recent research has developed driving-oriented pretraining methods for visual encoders~\citep{wu2023ppgeo,yang2024vidar,min2024driveworld,wang2026drivejepa,strong2026lfg}. These results show that driving benefits from representations beyond generic image recognition. However, visual structure, semantic context, and scene geometry are often learned through different objectives. Learning a comprehensive representation of driving scenes without increasing encoder size therefore remains a central challenge.

Beyond stronger features from individual images, driving representations should capture the surrounding scene across cameras and its evolution over time. Recent work has shown that multi-camera observations and historical frames can improve trajectory planning~\citep{hu2022stp3,hu2023uniad,xie2025s4driver}. However, encoding each image separately does not model spatiotemporal relationships. Dense processing of every camera frame also increases inference latency. A further challenge is to adapt a visual encoder to provide spatiotemporal representations for planning while limiting
the cost of processing multiple cameras and frames.

To address these challenges, we propose ConDenseDrive, a three-stage framework that makes a visual encoder knowledge-dense and token-sparse. First, multi-teacher visual distillation transfers complementary visual, semantic, and geometric knowledge into the encoder, improving representation quality without increasing encoder size. Second, multi-frame register adaptation uses the distilled encoder to capture spatiotemporal context in register tokens for planning. Finally, register-guided token compression uses these driving-oriented register tokens to retain informative patch tokens and fuse the remaining content, reducing the visual encoding cost of multiple cameras and frames without an auxiliary routing network.

ConDenseDrive achieves 94.5 PDMS on NAVSIM v1 and 90.8 EPDMS on NAVSIM v2, outperforming strong approaches centered on scene representation and planner design. It further reaches 37.4 HD-Score in zero-shot closed-loop evaluation on HUGSIM. These gains arise from encoder-centered optimization without redesigning the downstream planner. Its knowledge-dense and token-sparse design delivers a $2.0\times$ speedup while improving planning performance. These findings demonstrate that visual encoder design provides an effective route to reliable and efficient end-to-end driving.

Our main contributions are summarized as follows:
\begin{itemize}
    \item We introduce ConDenseDrive, a three-stage framework that makes the visual encoder knowledge-dense and token-sparse for end-to-end driving.
    \item We propose register-guided token compression, which uses driving-oriented register tokens to retain informative patch tokens without an auxiliary routing network.
    \item We achieve state-of-the-art performance on NAVSIM v1 and v2 with a $2.0\times$ end-to-end speedup and demonstrate zero-shot closed-loop generalization on HUGSIM.
\end{itemize}

\section{Related Work}
\label{sec:related_work}

\textbf{End-to-End Autonomous Driving.} Recent progress has largely come from redesigning scene representations and planners. UniAD unifies perception, prediction, and planning through a BEV representation, while VAD represents the scene with vectorized elements~\citep{hu2023uniad,jiang2023vad}. Planning has evolved from trajectory vocabularies in VADv2 to generative distributions in DiffusionDrive and learned proposal selection in DriveSuprim~\citep{jiang2026vadv2,liao2025diffusiondrive,yao2025drivesuprim}. VLA systems further introduce semantic reasoning into end-to-end driving. ReCogDrive transfers VLM-derived driving knowledge to a diffusion planner, while UniDriveVLA coordinates specialized experts for understanding, perception, and action planning~\citep{li2025recogdrive,li2026unidrivevla}. RAD and RAD-2 instead improve driving policies through large-scale closed-loop reinforcement learning~\citep{gao2025rad,gao2026rad2}. DrivoR connects a ViT encoder to lightweight planning modules through camera-specific register tokens, but still processes all patch tokens within each camera image~\citep{kirby2026drivor}. We improve representation quality while reducing patch token computation inside the encoder.

\textbf{Visual Knowledge Distillation.} Knowledge distillation transfers capabilities from large teachers to compact students~\citep{hinton2015distilling,touvron2021deit}. Recent methods extend feature distillation across diverse vision foundation models. AM-RADIO and UNIC consolidate complementary visual teachers within a shared encoder, while DUNE aligns representations from 2D and 3D teachers~\citep{ranzinger2024amradio,sariyildiz2024unic,sariyildiz2025dune}. Theia distills diverse vision foundation models into a compact encoder for robot policy learning~\citep{shang2025theia}. DriveZero further applies multi-teacher visual distillation to learn a visual backbone for end-to-end driving~\citep{he2026drivezero}. We jointly address visual knowledge transfer and encoder efficiency for end-to-end driving.

\textbf{Token Compression.} Token compression reduces the cost of Vision Transformers by changing how patch tokens are processed~\citep{dosovitskiy2021vit}. DynamicViT learns to prune less informative tokens, while A-ViT allocates a different computation depth to each token~\citep{rao2021dynamicvit,yin2022avit}. EViT ranks patch tokens using CLS attention and fuses the low-score tokens into a single representation~\citep{liang2022evit}. ToMe merges similar tokens, while TPS combines token pruning and squeezing to preserve discarded information~\citep{bolya2023tome,wei2023tps}. Driving-specific methods operate on different compression targets. SeeLess stochastically masks extracted foundation-model patch features, while COMPACT-VA compresses long temporal context into a bounded working memory~\citep{mallak2026seeless,liang2026compactva}. Our method instead uses attention from driving-oriented register tokens to retain informative patch tokens and fuse the remaining low-score tokens into a single token within selected encoder blocks.

\section{Preliminaries}
\label{sec:preliminaries}

\noindent\textbf{Problem Definition.}\quad
Let $\mathcal{I}=\{I_{t,v}\}$ denote the set of synchronized images from $V$ camera views over $T$ observed frames, where $t$ and $v$ index time and camera view. Let $\mathbf{u}$ denote the auxiliary inputs, including the ego state and navigation information. The predicted trajectory $\boldsymbol{\tau}=\{(x_{q},y_{q},\psi_{q})\}_{q=1}^{Q}$ contains $Q$ future ego poses. We formulate the task as
\begin{equation}
    \boldsymbol{\tau}
    =\mathcal{P}\!\left(\mathcal{E}(\mathcal{I}),\mathbf{u}\right),
    \label{eq:e2e_task}
\end{equation}
where \(\mathcal E\) denotes the visual encoding module and $\mathcal{P}$ denotes the downstream planner. We optimize $\mathcal{E}$ while retaining the interface of $\mathcal{P}$. Since every view and frame contributes a patch token sequence, visual encoding is costly. Our goal is to learn driving-oriented spatiotemporal representations while reducing computation on less relevant patch tokens without degrading trajectory prediction.

\section{Method}
\label{sec:method}

\begin{figure}[t]
    \centering
    \includegraphics[width=\linewidth]{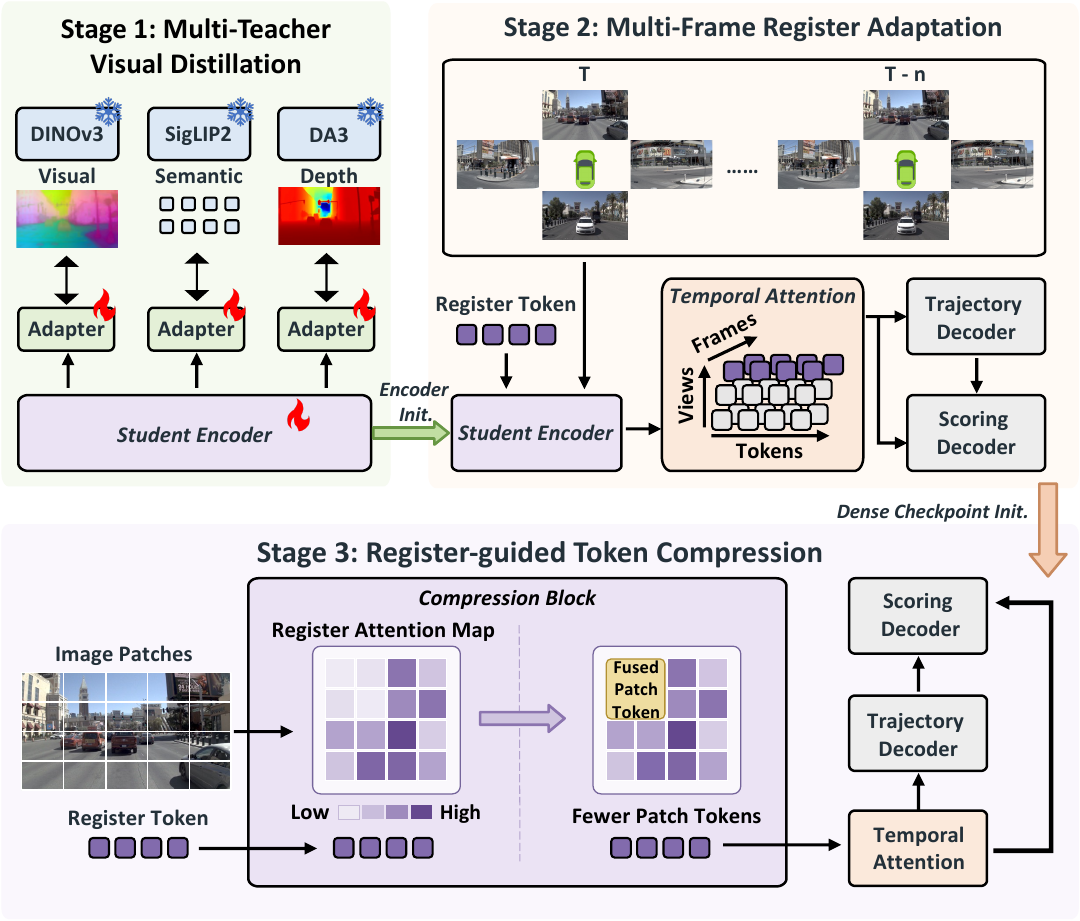}
    \caption{\textbf{Three-stage framework of ConDenseDrive.}
    Multi-teacher visual distillation enriches the student encoder. Multi-frame register adaptation incorporates spatiotemporal context into register tokens for planning. Register-guided token compression then uses the resulting register-to-patch attention to retain informative patch tokens and fuse the rest.}
    \label{fig:overview}
\end{figure}

\noindent\textbf{Overview.}\quad
Fig.~\ref{fig:overview} presents the three-stage framework of ConDenseDrive. First, multi-teacher visual distillation enriches the representations of a compact encoder without increasing its size. Second, multi-frame register adaptation uses the distilled encoder to aggregate spatial evidence and temporal context in register tokens for planning. Finally, register-guided token compression uses attention from these register tokens to retain informative patch tokens and fuse the rest, reducing visual encoding cost.

\subsection{Multi-Teacher Visual Distillation}
End-to-end driving relies on diverse visual cues, including appearance, semantic context, and scene geometry. Vision foundation models trained with distinct objectives encode these cues in complementary ways. To make a compact encoder knowledge-dense without increasing its size, Stage 1 distills their features into a single student encoder.

\noindent\textbf{Teacher Supervision.}\quad
We use DINOv3~\citep{simeoni2025dinov3}, SigLIP2~\citep{tschannen2025siglip2}, and Depth Anything~3 (DA3)~\citep{lin2026depthanything3} as frozen teachers for visual, semantic, and geometric supervision, respectively.
We extract final-layer patch features from DINOv3 and SigLIP2, and the final DA3 backbone output before its depth decoder. 

\noindent\textbf{Feature Alignment.}\quad
The teacher features differ in spatial resolution, channel width, and scale. Given an image \(I\), the DeiT-S student encoder~\citep{touvron2021deit} produces patch features \(\mathbf F_s\). For each teacher \(m\), we normalize its features per channel to obtain \(\bar{\mathbf F}_m\), while a teacher-specific convolutional adapter \(A_m\) maps \(\mathbf F_s\) to match \(\bar{\mathbf F}_m\) in spatial resolution and channel width. All branches supervise dense patch features only, while CLS token features are excluded.

\noindent\textbf{Distillation Objective.}\quad
Once aligned, each adapted student feature \(A_m(\mathbf F_s)\) is matched to its normalized teacher target \(\bar{\mathbf F}_m\) using cosine and Smooth L1 losses. With $M$ teachers and loss weights $\lambda_{\mathrm{cos}}$ and $\lambda_{\mathrm{L1}}$, each branch receives equal weight:
\begin{equation}
    \mathcal{L}_{\mathrm{distill}}
    =\frac{1}{M}\sum_{m=1}^{M}\big[
    \lambda_{\mathrm{cos}}\mathcal{L}_{\mathrm{cos}}
    \!\left(A_m(\mathbf{F}_s),\bar{\mathbf{F}}_m\right)
    +\lambda_{\mathrm{L1}}\mathcal{L}_{\mathrm{SmoothL1}}
    \!\left(A_m(\mathbf{F}_s),\bar{\mathbf{F}}_m\right)\big].
    \label{eq:distillation_loss}
\end{equation}
The cosine term aligns feature direction, while Smooth L1 encourages element-wise agreement. Only the student and adapters are optimized in this stage. After distillation, we discard the adapters and use the student encoder to initialize Stage 2, where register tokens aggregate its enriched patch features for planning.

\subsection{Multi-Frame Register Adaptation}
Multi-teacher visual distillation enriches the encoder's image features, whereas planning requires spatiotemporal representations across camera views and frames. Stage 2 places the distilled encoder within the planning architecture of DrivoR~\citep{kirby2026drivor} and adapts it through end-to-end training. All patch tokens are retained during image encoding at this stage.

\noindent\textbf{Spatial Attention.}\quad
For each image $I_{t,v}$, the shared encoder processes camera-specific register tokens together with its patch tokens. Through self-attention, the registers attend to patches across the image and aggregate spatial information into $\mathbf R_{t,v}$.

\noindent\textbf{Temporal Attention.}\quad
Each image is initially encoded independently, so its register tokens do not interact across time. To capture scene evolution, we arrange the frame-wise register tokens of each camera into one sequence and add learned temporal embeddings and ego-motion embeddings derived from relative planar motion. We denote the combined embeddings by $\mathbf{E}_v$:
\begin{equation}
    \mathbf{R}_{v}
    =\operatorname{TempAttn}
    \!\left([\mathbf{R}_{1,v};\ldots;\mathbf{R}_{T,v}]+\mathbf{E}_v\right).
\label{eq:temporal_registers}
\end{equation}
Temporal attention operates independently within each camera stream. The register tokens from all views and observed frames are passed to the trajectory generator and scorer together with the ego state and navigation information $\mathbf{u}$. 

\noindent\textbf{End-to-End Objective.}\quad
We retain the planner's trajectory generation and scoring objectives during end-to-end training:
\begin{equation}
    \mathcal{L}_{\mathrm{E2E}}
    =\mathcal{L}_{\mathrm{traj}}+\mathcal{L}_{\mathrm{score}},
    \label{eq:e2e_loss}
\end{equation}
where $\mathcal{L}_{\mathrm{traj}}$ is the minimum L1 loss over the trajectory candidates and $\mathcal{L}_{\mathrm{score}}$ is the binary cross-entropy loss for driving-quality scores. This objective adapts the spatiotemporal register representations to planning and shapes the register-to-patch attention used for token routing in Stage 3.

\subsection{Register-Guided Token Compression}
Multi-camera, multi-frame inputs provide richer visual evidence but increase the number of patch tokens processed by the encoder. Stage 3 therefore uses attention from the registers adapted for planning to reduce redundant patch computation. It starts from the complete Stage 2 checkpoint while retaining the planner and end-to-end objective.

\noindent\textbf{Register-Guided Routing.}\quad
We apply token compression in selected Transformer blocks. At each compression block, let \(\mathbf{x}_1,\ldots,\mathbf{x}_N\) denote the compressible sequence, consisting of unfused patch tokens and fused tokens carried forward from earlier blocks. We score each token by averaging the attention it receives from $R$ register tokens over $H$ attention heads:
\begin{equation}
    s_i
    =\frac{1}{HR}
    \sum_{h=1}^{H}\sum_{j=1}^{R}
    w_{h,j,i},
    \label{eq:register_routing_score}
\end{equation}
where $w_{h,j,i}$ is the attention weight from register token $j$ to token $\mathbf{x}_i$ in head $h$. A higher score indicates stronger attention from the registers trained for planning. No auxiliary routing network or loss is introduced. The register attention used to compute the routing scores is shaped by the end-to-end driving objective in Equation~\eqref{eq:e2e_loss}.

\noindent\textbf{Token Retention and Fusion.}\quad
We always preserve the register and CLS tokens. At each compression block, among the current compressible tokens, we retain the fraction $\rho$ with the highest routing scores and merge all lower-scored tokens into a single fused token $\mathbf{x}_{\mathrm{fuse}}$. We compute it as $\mathbf{x}_{\mathrm{fuse}}=\sum_i s_i\mathbf{x}_i$, where the sum is taken over the lower-scored tokens. The routing scores are not renormalized, preserving the original attention mass of the lower-scored set.  The fused token proceeds through subsequent blocks and remains eligible for later compression.

\begin{table}[t]
\caption{\textbf{Comparison on NAVSIM v1 \texttt{navtest} split.} ConDenseDrive uses $T=2$ and $\rho=0.4$. Bold values denote the best result among the listed learned methods.}
\label{tab:navsim-v1-results}
\centering
\begingroup
\setlength{\tabcolsep}{2.5pt}
\resizebox{\linewidth}{!}{%
\begin{tabular}{l|l|ccccc|>{\columncolor{scoregraylight}}c}
\toprule
\textbf{Method} & \textbf{Image Encoder} & \textbf{NC}$\uparrow$ & \textbf{DAC}$\uparrow$ & \textbf{TTC}$\uparrow$ & \textbf{Comf.}$\uparrow$ & \textbf{EP}$\uparrow$ & \textbf{PDMS}$\uparrow$ \\
\midrule
Human driver~\citep{dauner2024navsim}
& -- & 100 & 100 & 100 & 99.9 & 87.5 & 94.8 \\
\midrule
UniAD~\citep{hu2023uniad}
& ResNet-50 & 97.8 & 91.9 & 92.9 & \textbf{100} & 78.8 & 83.4 \\
TransFuser~\citep{chitta2023transfuser}
& ResNet-34 & 97.7 & 92.8 & 92.8 & \textbf{100} & 79.2 & 84.0 \\
DiffusionDrive~\citep{liao2025diffusiondrive}
& ResNet-34 & 98.2 & 96.2 & 94.7 & \textbf{100} & 82.2 & 88.1 \\
Hydra-MDP~\citep{li2024hydramdp}
& V2-99 & 98.4 & 97.8 & 93.9 & \textbf{100} & 86.5 & 90.3 \\
GoalFlow~\citep{xing2025goalflow}
& V2-99 & 98.4 & 98.3 & 94.6 & \textbf{100} & 85.0 & 90.3 \\
DiffusionDriveV2~\citep{zou2025diffusiondrivev2}
& ResNet-34 & 98.3 & 97.9 & 94.8 & 99.9 & 87.5 & 91.2 \\
DriveSuprim~\citep{yao2025drivesuprim}
& ViT-L & 98.6 & 98.6 & 95.5 & \textbf{100} & 91.3 & 93.5 \\
DrivoR~\citep{kirby2026drivor}
& DINOv2-Small & 99.0 & \textbf{98.9} & 96.7 & \textbf{100} & 90.0 & 93.7 \\
RAP-DINO~\citep{feng2025rap}
& DINOv3-H+ & 99.1 & \textbf{98.9} & 96.7 & \textbf{100} & 90.3 & 93.8 \\
\midrule
\textbf{ConDenseDrive (ours)}
& \textbf{Distilled DeiT-S} & \textbf{99.3} & 98.8 & \textbf{96.8} & \textbf{100} & \textbf{91.9} & \textbf{94.5} \\
\bottomrule
\end{tabular}%
}
\endgroup

\caption{\textbf{Comparison on NAVSIM v2 \texttt{navtest} split.} ConDenseDrive uses $T=2$ and $\rho=0.4$. EPDMS$^{*}$ and EPDMS denote scores before and after the official evaluator correction, respectively.}
\label{tab:navsim-v2-results}
\centering
\begingroup
\setlength{\tabcolsep}{2.2pt}
\renewcommand{\arraystretch}{0.98}
\resizebox{\linewidth}{!}{%
\begin{tabular}{l|ccccccccc|>{\columncolor{scoregraylight}}c>{\columncolor{scoregraydark}}c}
\toprule
\textbf{Method}
& \textbf{NC}$\uparrow$
& \textbf{DAC}$\uparrow$
& \textbf{DDC}$\uparrow$
& \textbf{TLC}$\uparrow$
& \textbf{EP}$\uparrow$
& \textbf{TTC}$\uparrow$
& \textbf{LK}$\uparrow$
& \textbf{HC}$\uparrow$
& \textbf{EC}$\uparrow$
& \textbf{EPDMS}$^{*}\uparrow$
& \textbf{EPDMS}$\uparrow$ \\
\midrule
TransFuser~\citep{chitta2023transfuser}
& 96.9 & 89.9 & 97.8 & 99.7 & 87.1 & 95.4 & 92.7 & 98.3 & 87.2 & 76.7 & -- \\
DiffusionDriveV2~\citep{zou2025diffusiondrivev2}
& 97.7 & 96.6 & 99.2 & \textbf{99.8} & 88.9 & 97.2 & 96.0 & 97.8 & \textbf{91.0} & 85.5 & 87.5 \\
Hydra-MDP++~\citep{li2025hydramdpplus}
& 98.4 & 98.0 & 99.4 & \textbf{99.8} & 87.5 & 97.7 & 95.3 & 98.3 & 77.4 & 85.1 & -- \\
Senna-2~\citep{song2026senna2}
& 98.5 & 97.8 & 99.5 & \textbf{99.8} & 88.1 & 97.5 & \textbf{97.0} & \textbf{98.6} & 88.4 & 86.6 & -- \\
SparseDriveV2~\citep{sun2026sparsedrivev2}
& 98.1 & 98.1 & 99.6 & \textbf{99.8} & \textbf{91.1} & 97.3 & 96.9 & 98.2 & 78.4 & 86.7 & 90.1 \\
DriveSuprim~\citep{yao2025drivesuprim}
& 98.4 & 98.6 & 99.6 & \textbf{99.8} & 90.5 & 97.8 & \textbf{97.0} & 98.3 & 78.6 & 87.1 & -- \\
CLOVER~\citep{ang2026clover}
& \textbf{99.4} & \textbf{99.3} & 99.5 & \textbf{99.8} & 86.9 & \textbf{98.9} & 95.2 & 98.2 & 75.5 & \textbf{87.2} & 90.4 \\
\midrule
\textbf{ConDenseDrive (ours)}
& \textbf{99.4} & 99.1 & \textbf{99.7} & 99.5 & 90.3 & 98.8 & 96.1 & 97.8 & 71.0 & 87.1 & \textbf{90.8} \\
\bottomrule
\end{tabular}%
}
\endgroup

\end{table}

\begin{figure}[t]
    \centering
    \begin{minipage}[t]{0.49\linewidth}
        \centering
        \includegraphics[width=\linewidth]{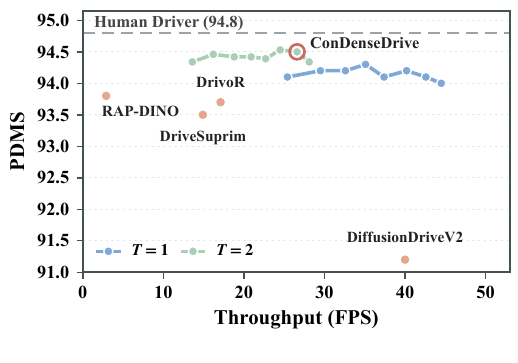}
        \vspace{1pt}
        \textbf{(a) Accuracy--efficiency trade-off.}
    \end{minipage}\hfill
    \begin{minipage}[t]{0.49\linewidth}
        \centering
        \includegraphics[width=\linewidth]{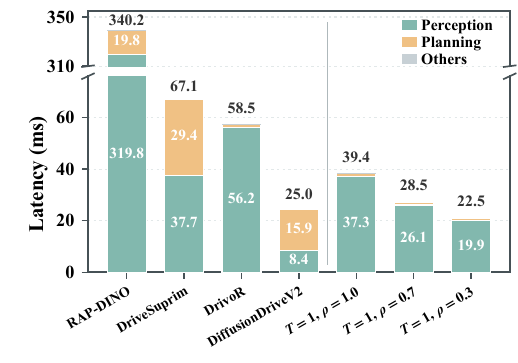}
        \vspace{1pt}
        \textbf{(b) Module-wise latency.}
    \end{minipage}
    \caption{\textbf{Accuracy--efficiency trade-off and module-wise latency.}
    (a) PDMS versus throughput for $T=1$ and $T=2$. Markers follow $\rho=1.0,0.9,\ldots,0.3$ from left to right, and the outlined marker denotes the selected $T=2$, $\rho=0.4$ model.
    (b) End-to-end latency decomposed into perception, planning, and other costs.}
    \label{fig:efficiency-tradeoff}
    \par\vspace{0.35em}
    \begingroup
    \makeatletter\def\@captype{table}\makeatother
    \renewenvironment{table*}[1][]{\par}{\par}
    \begin{table*}[t]
\centering
\caption{\textbf{Zero-shot closed-loop evaluation on HUGSIM.}
ConDenseDrive uses $T=1$ and $\rho=0.7$. }
\label{tab:hugsim-436}
\begingroup
\setlength{\tabcolsep}{3.0pt}
\renewcommand{\arraystretch}{1.08}
\resizebox{\textwidth}{!}{%
\begin{tabular}{l|cccccc|cccc>{\columncolor{scoregraylight}}c}
\toprule
\multirow{2}{*}{\textbf{Method}}
& \multicolumn{6}{c|}{\textbf{All 436 Scenarios}}
& \multicolumn{5}{c}{\textbf{HD-Score}} \\
\cmidrule(lr){2-7}
\cmidrule(lr){8-12}
& \textbf{NC}$\uparrow$
& \textbf{DAC}$\uparrow$
& \textbf{TTC}$\uparrow$
& \textbf{COM}$\uparrow$
& \textbf{PDMS}$\uparrow$
& \textbf{RC}$\uparrow$
& \textbf{Easy}$\uparrow$
& \textbf{Medium}$\uparrow$
& \textbf{Hard}$\uparrow$
& \textbf{Extreme}$\uparrow$
& \textbf{Average}$\uparrow$ \\
\midrule
VAD~\citep{jiang2023vad}
& 41.2 & 90.3 & 28.0 & \textbf{95.3} & 28.3 & 30.1
& 42.0 & 8.5 & 7.7 & 6.3 & 13.9 \\
UniAD~\citep{hu2023uniad}
& \textbf{65.6} & 93.2 & 51.6 & 66.3 & 49.4 & 43.8
& 64.0 & 37.2 & 21.0 & 6.3 & 31.2 \\
LTF~\citep{chitta2023transfuser}
& 44.3 & 92.8 & 37.5 & 94.8 & 36.5 & 38.0
& 66.1 & 15.5 & 12.0 & 11.7 & 23.1 \\
DrivoR~\citep{kirby2026drivor}
& 52.2 & 95.6 & 46.2 & 93.9 & 44.8 & 47.2
& 78.0 & 29.1 & 20.0 & \textbf{14.1} & 32.5 \\
\midrule
\textbf{ConDenseDrive (ours)}
& 60.3 & \textbf{96.8} & \textbf{54.1} & 93.9
& \textbf{53.3} & \textbf{49.9}
& \textbf{79.7} & \textbf{41.2}
& \textbf{21.8} & 13.3 & \textbf{37.4} \\
\bottomrule
\end{tabular}%
}
\endgroup
\end{table*}

    \endgroup
\end{figure}

\section{Experiments}
\label{sec:experiments}

\subsection{Experimental Setup}
\textbf{Benchmarks and Metrics.}
We train on the NAVSIM \texttt{navtrain} split with $103{,}288$ scenes and evaluate on \texttt{navtest} with $12{,}146$ scenes, using PDMS for NAVSIM v1 and EPDMS for NAVSIM v2~\citep{dauner2024navsim,cao2025pseudosimulation}. We further evaluate zero-shot closed-loop generalization on all 436 HUGSIM scenarios using HD-Score~\citep{zhou2024hugsim}.

\textbf{Implementation Details.}
We distill a DeiT-S student on front-view images from NAVSIM \texttt{navtrain} using frozen DINOv3 ViT-L/16, SigLIP2-SO400M NaFlex, and Depth Anything~3-Large teachers. End-to-end training uses front, rear, left, and right camera images resized to \(1136\times672\), with \(R=16\) register tokens per view per frame. The distilled backbone weights remain frozen and are adapted with LoRA~\citep{hu2022lora}, while the downstream planner is trained jointly. Training uses four A100 GPUs, and inference latency is measured on one. Further training and evaluation details are in Appendix~\ref{app:implementation-details}.

\subsection{Main Results}
\noindent\textbf{Result on NAVSIM v1.}\quad
ConDenseDrive achieves $94.5$ PDMS with $T=2$ and $\rho=0.4$, remaining only $0.3$ below the human-driver reference (Tab.~\ref{tab:navsim-v1-results}). Its NC, TTC, and EP scores of $99.3$, $96.8$, and $91.9$ indicate strong collision avoidance while maintaining ego progress. With a $22$M-parameter DeiT-S encoder, ConDenseDrive outperforms GoalFlow (V2-99, $70$M), RAP-DINO (DINOv3-H+, $840$M), and DriveSuprim (ViT-L, $304$M)~\citep{xing2025goalflow,feng2025rap,yao2025drivesuprim}. DriveZero also uses a multi-teacher distilled ViT-S encoder with a DrivoR-style planner, but relies on broader distillation data and additional RL trajectory supervision~\citep{he2026drivezero}. ConDenseDrive achieves higher PDMS through encoder optimization with less distillation data.
These comparisons show that stronger visual representations can improve planning without increasing encoder capacity or redesigning the downstream planner.

\noindent\textbf{Result on NAVSIM v2.}\quad
Using the same $T=2$ and $\rho=0.4$ configuration, ConDenseDrive achieves $90.8$ EPDMS with the updated official NAVSIM v2 evaluator and $87.1$ EPDMS$^{*}$ with the original evaluation code (Tab.~\ref{tab:navsim-v2-results}). CLOVER provides a relevant architectural comparison because it adopts a comparable DrivoR-style planner while focusing on proposal generation and trajectory ranking~\citep{ang2026clover}. ConDenseDrive instead focuses on visual encoder design and outperforms CLOVER by $0.4$ EPDMS. It also improves EP by $3.4$ points while matching CLOVER in NC and maintaining a comparable TTC score. This result supports visual encoder optimization as another effective direction for improving end-to-end planning.

\noindent\textbf{Zero-Shot on HUGSIM.}\quad
Our three-stage training pipeline uses no data from any of HUGSIM’s four source datasets. Under this zero-shot setting, ConDenseDrive with $T=1$ and $\rho=0.7$ achieves $37.4$ HD-Score on all 436 scenarios, outperforming DrivoR by $4.9$ points (Tab.~\ref{tab:hugsim-436}). The improvement is particularly pronounced on medium-difficulty scenarios, where the HD-Score increases from $29.1$ to $41.2$. NC, TTC, and RC also improve over DrivoR, indicating gains in collision avoidance and route completion. Since the downstream planner remains unchanged, this gain is achieved through encoder-centered optimization.

\noindent\textbf{Inference Efficiency.}\quad
Fig.~\ref{fig:efficiency-tradeoff}(a) compares driving performance and throughput under the same hardware setting. Our $T=2$, $\rho=0.4$ model achieves $94.5$ PDMS at $26.6$ FPS. At $T=2$, reducing $\rho$ from $1.0$ to $0.4$ provides a $2.0\times$ speedup while improving driving performance. ConDenseDrive is also faster and more accurate than RAP-DINO, DriveSuprim, and DrivoR. DiffusionDriveV2 is faster than the selected two-frame model but scores $3.3$ points lower in PDMS. With $T=1$ and $\rho=0.3$, ConDenseDrive reaches $44.5$ FPS at $94.0$ PDMS, only $0.1$ below its uncompressed counterpart. Across the $T=1$ and $T=2$ settings, throughput increases as $\rho$ decreases, while PDMS varies by no more than $0.3$. This stability suggests that a substantial share of dense patch-token computation can be avoided.

Fig.~\ref{fig:efficiency-tradeoff}(b) shows that visual encoding dominates latency in most end-to-end models. For ViT-based systems, dense patch-token sequences are processed across successive Transformer blocks for each camera view and input frame, amplifying this latency bottleneck. At $T=1$, token compression reduces perception latency by $30.0\%$ at $\rho=0.7$ and by $46.6\%$ at $\rho=0.3$, relative to $\rho=1.0$. These reductions show that optimizing the visual encoder alone can substantially increase end-to-end system throughput.

\subsection{Ablation Studies}
We evaluate the complete three-stage pipeline to quantify how each stage affects planning performance and inference efficiency. We then isolate the key design choices in multi-teacher distillation, temporal modeling, and register-guided token compression.

\begin{table}[!ht]
\centering
\par\noindent
\begin{minipage}{\linewidth}
\makeatletter\def\@captype{table}\makeatother
\centering
\caption{\textbf{Contribution of the three-stage framework.}
Distill., Temporal, and Compress. denote visual distillation, temporal attention, and token compression, respectively. ID~1 uses the original pretrained DeiT-S.}
\label{tab:stage-contribution}
\centering
\begingroup
\setlength{\tabcolsep}{2.4pt}
\resizebox{\linewidth}{!}{%
\begin{tabular}{c|c|ccc|cc|c|ccccc|>{\columncolor{scoregraylight}}c}
\toprule
\textbf{ID}
& \textbf{Encoder}
& \textbf{Distill.}
& \textbf{Temporal}
& \textbf{Compress.}
& \textbf{Frames}
& \textbf{Rate}
& \textbf{FPS}$\uparrow$
& \textbf{NC}$\uparrow$
& \textbf{DAC}$\uparrow$
& \textbf{TTC}$\uparrow$
& \textbf{Comf.}$\uparrow$
& \textbf{EP}$\uparrow$
& \textbf{PDMS}$\uparrow$ \\
\midrule
1
& \multirow{4}{*}{DeiT-S} & $\times$ & $\times$ & $\times$ & $T=1$ & $\rho=1.0$ & 25.4
& 98.7 & 97.9 & 95.6 & \textbf{100} & 88.0 & 91.8$_{\phantom{\mathbf{+2.3}}}$ \\
2
& & $\surd$ & $\times$ & $\times$ & $T=1$ & $\rho=1.0$ & 25.4
& 99.1 & 98.8 & 96.6 & \textbf{100} & 91.2 & 94.1$_{\textcolor{scoregreen}{\mathbf{+2.3}}}$ \\
3
& & $\surd$ & $\surd$ & $\times$ & $T=2$ & $\rho=1.0$ & 13.6
& 99.2 & \textbf{98.9} & \textbf{97.0} & \textbf{100} & 91.2 & 94.3$_{\textcolor{scoregreen}{\mathbf{+2.5}}}$ \\
4
& & $\surd$ & $\surd$ & $\surd$ & $T=2$ & $\rho=0.4$ & \textbf{26.6}
& \textbf{99.3} & 98.8 & 96.8 & \textbf{100} & \textbf{91.9} & \textbf{94.5}$_{\textcolor{scoregreen}{\mathbf{+2.7}}}$ \\
\bottomrule
\end{tabular}
}
\endgroup

\vspace{-0.5em}
\begin{minipage}[t]{0.495\linewidth}
\centering
\caption{\textbf{Routing token ($T=1$, $\rho=0.7$).}}
\label{tab:routing-token-ablation}
\begingroup
\fontsize{6.8pt}{8.1pt}\selectfont
\setlength{\tabcolsep}{0.7pt}
\renewcommand{\arraystretch}{1.10}
\begin{tabular*}{\linewidth}{l|@{\extracolsep{0pt}\hspace{2.2pt}}c@{\extracolsep{\fill}}cccc@{\extracolsep{0pt}\hspace{2.2pt}}|>{\columncolor{scoregraylight}[\tabcolsep][0pt]\hspace{1.5pt}}c<{\hspace{1.5pt}}@{}}
\toprule
\textbf{Routing Token}
& \textbf{NC}$\uparrow$
& \textbf{DAC}$\uparrow$
& \textbf{TTC}$\uparrow$
& \textbf{Comf.}$\uparrow$
& \textbf{EP}$\uparrow$
& \textbf{PDMS}$\uparrow$ \\
\midrule
CLS & \textbf{98.9} & 98.9 & 96.1 & \textbf{100} & 91.5 & 94.0 \\
Register & \textbf{98.9} & \textbf{99.0} & \textbf{96.3} & \textbf{100} & \textbf{91.8} & \textbf{94.3} \\
\bottomrule
\end{tabular*}
\endgroup
\end{minipage}\hfill
\begin{minipage}[t]{0.495\linewidth}
\centering
\caption{\textbf{Compression init. ($T=1$, $\rho=0.7$).}}
\label{tab:initialization-ablation}
\begingroup
\fontsize{6.8pt}{8.1pt}\selectfont
\setlength{\tabcolsep}{0.7pt}
\renewcommand{\arraystretch}{1.10}
\begin{tabular*}{\linewidth}{l|@{\extracolsep{0pt}\hspace{2.2pt}}c@{\extracolsep{\fill}}cccc@{\extracolsep{0pt}\hspace{2.2pt}}|>{\columncolor{scoregraylight}[\tabcolsep][0pt]\hspace{1.5pt}}c<{\hspace{1.5pt}}@{}}
\toprule
\textbf{Initialization}
& \textbf{NC}$\uparrow$
& \textbf{DAC}$\uparrow$
& \textbf{TTC}$\uparrow$
& \textbf{Comf.}$\uparrow$
& \textbf{EP}$\uparrow$
& \textbf{PDMS}$\uparrow$ \\
\midrule
Scratch & \textbf{98.9} & 98.7 & \textbf{96.3} & \textbf{100} & 90.1 & 93.6 \\
Dense checkpoint & \textbf{98.9} & \textbf{99.0} & \textbf{96.3} & \textbf{100} & \textbf{91.8} & \textbf{94.3} \\
\bottomrule
\end{tabular*}
\endgroup
\end{minipage}
\end{minipage}
\par

\end{table}

\begin{wraptable}{r}{0.48\textwidth}
\vspace{-0.35em}
\centering
\centering
\caption{\textbf{Keep rate ($T=2$).}}
\label{tab:keep-rate-ablation}
\begingroup
\setlength{\tabcolsep}{5.0pt}
\renewcommand{\arraystretch}{1.16}
\resizebox{\linewidth}{!}{%
\begin{tabular}{c|cc|>{\columncolor{scoregraylight}}c}
\toprule
\textbf{Rate}
& \textbf{Encoder GFLOPs}$\downarrow$
& \textbf{FPS}$\uparrow$
& \textbf{PDMS}$\uparrow$ \\
\midrule
$\rho=1.0$ & 2363.62 & 13.6 & 94.3 \\
$\rho=0.9$ & 1951.77 & 16.2 & 94.5 \\
$\rho=0.8$ & 1637.24 & 18.8 & 94.4 \\
$\rho=0.7$ & 1398.12 & 20.9 & 94.4 \\
$\rho=0.6$ & 1215.52 & 22.7 & 94.4 \\
$\rho=0.5$ & 1075.51 & 24.5 & 94.5 \\
$\rho=0.4$ & 968.67 & 26.6 & \textbf{94.5} \\
$\rho=0.3$ & \textbf{887.02} & \textbf{28.1} & 94.3 \\
\bottomrule
\end{tabular}%
}
\endgroup

\vspace*{-0.1em}
\end{wraptable}

\noindent\textbf{Contribution of Each Stage.}\quad
Tab.~\ref{tab:stage-contribution} separates the contribution of each stage. ID 1 directly uses the original pretrained DeiT-S encoder and reaches $91.8$ PDMS. Multi-teacher distillation increases PDMS by $2.3$ points to $94.1$ without changing throughput. Extending the input to two frames with temporal attention increases PDMS from $94.1$ to $94.3$, although using two frames reduces throughput to $13.6$ FPS. Register-guided token compression at $\rho=0.4$ restores throughput to $26.6$ FPS and reaches $94.5$ PDMS. 

\noindent\textbf{Routing Token.}\quad
With $T=1$ and $\rho=0.7$ fixed, routing with the CLS token reaches $94.0$ PDMS, while routing with register tokens reaches $94.3$ (Tab.~\ref{tab:routing-token-ablation}). Since register tokens form the visual interface to the planner, their attention is more directly aligned with planning-relevant information.

\noindent\textbf{Compression Initialization.}\quad
Tab.~\ref{tab:initialization-ablation} compares two initialization strategies with $T=1$ and $\rho=0.7$. Scratch initialization loads the Stage 1 checkpoint only for the visual encoder and initializes the remaining components from scratch. Dense-checkpoint initialization loads all weights from Stage 2 before enabling token compression. This increases PDMS from $93.6$ to $94.3$, showing that driving adaptation before compression provides a stronger initialization.

\noindent\textbf{Keep Rate.}\quad
Tab.~\ref{tab:keep-rate-ablation} evaluates different keep rates with $T=2$. All compressed variants use the same training hyperparameters without rate-specific tuning. At $\rho=0.4$, encoder computation decreases by $59.0\%$ and throughput reaches $26.6$ FPS, while PDMS improves from $94.3$ to $94.5$. Reducing $\rho$ to $0.3$ further raises throughput to $28.1$ FPS while matching the $\rho=1.0$ result. These results show that patch token computation can be substantially reduced without degrading driving performance.

\begin{figure}[t]
    \centering
    \includegraphics[width=\linewidth]{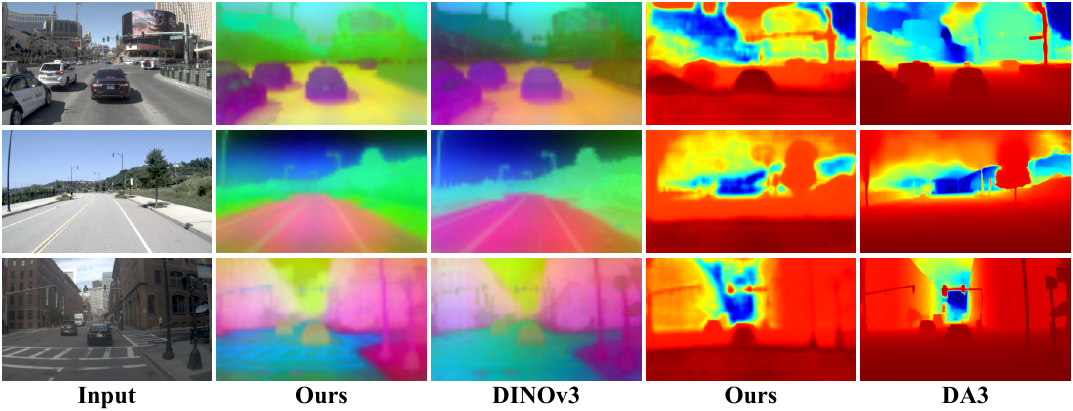}
    \caption{\textbf{PCA and depth visualizations.}
    From left to right, columns show the input image, PCA projections of the DeiT-S student and DINOv3 ViT-L/16 teacher, and depth maps from the student and Depth Anything~3-Large teacher.}
    \label{fig:distilled-feature-vis}
    \par\vspace{0.45em}
    \includegraphics[width=\linewidth]{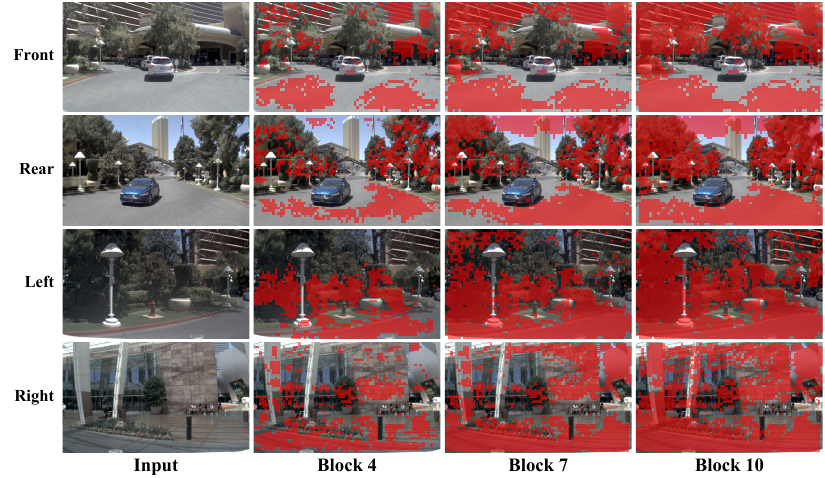}
    \caption{\textbf{Progressive register-guided token compression.} Rows show front, rear, left, and right camera views of the same scene. Columns show the input image and patch token states at Transformer blocks~4, 7, and 10 with $\rho=0.7$. Red regions denote patch tokens assigned to fusion, while unmasked regions indicate retained patch tokens.}
    \label{fig:token-compression-vis}
\end{figure}

\subsection{Qualitative Analysis}
\noindent\textbf{Feature Visualizations.}\quad
Fig.~\ref{fig:distilled-feature-vis} compares adapter-aligned student features with their teacher references across three scenes. Compared with DINOv3, the student PCA projections more clearly delineate small objects and fine-grained road surface structure, while separating vehicle windows from surrounding body regions that DINOv3 tends to merge. For depth visualization, we decode the DA3-aligned student features using a lightweight convolutional head because the original DA3 head aggregates multiple feature layers. The resulting depth maps preserve the broad scene geometry of the DA3 predictions. These results show that the shared student representation captures complementary appearance and geometric cues from multiple teachers.

\noindent\textbf{Token Compression Visualizations.}\quad
Fig.~\ref{fig:token-compression-vis} presents token compression patterns across the compression blocks. As compression proceeds, retained tokens increasingly concentrate on vehicles, pedestrians, and nearby obstacles, while tokens over the sky and distant buildings are fused. The allocation also varies across camera views. The front view preserves more road-surface tokens near the driving path, whereas side views prioritize nearby obstacles. These observations indicate that register-guided routing adjusts token retention according to the visual context of each view.

\section{Conclusion}
\label{sec:conclusion}
We present ConDenseDrive, which revisits the visual encoder as a central design target for reliable and efficient end-to-end driving. The key idea is to enrich a compact encoder with complementary visual knowledge and concentrate computation on planning-relevant image patches. Our three-stage framework realizes this idea through multi-teacher distillation, multi-frame register adaptation, and register-guided token compression. Extensive experiments and qualitative analyses demonstrate that ConDenseDrive improves planning performance and inference efficiency. We hope this work encourages further research on driving-oriented visual encoders.

\subsection*{AI use statement}

Large Language Models (LLMs) were used solely to assist with manuscript drafting, improve clarity, grammar, and readability, and identify relevant literature.

\bibliography{references}
\bibliographystyle{iclr2027_conference}

\clearpage
\appendix
\section{Implementation Details}
\label{app:implementation-details}

This section provides the evaluation metrics, three-stage training hyperparameters, and efficiency profiling protocol. Tab.~\ref{tab:training-configuration} summarizes the settings used for the main ConDenseDrive model.

\paragraph{NAVSIM evaluation.}
We use the official Predictive Driver Model Score (PDMS) for NAVSIM evaluation~\citep{dauner2024navsim}. PDMS measures safety, progress, and comfort by combining no-at-fault collisions (NC), drivable area compliance (DAC), time-to-collision (TTC), driving comfort (C), and ego progress (EP):
\begin{equation}
\mathrm{PDMS}
=
\mathrm{NC}\cdot\mathrm{DAC}\cdot
\frac{5\,\mathrm{EP}+5\,\mathrm{TTC}+2\,\mathrm{C}}{12}.
\label{eq:pdms}
\end{equation}

NAVSIM v2 adds driving direction compliance (DDC), traffic light compliance (TLC), lane keeping (LK), and extended comfort (EC), and replaces C with history comfort (HC)~\citep{cao2025pseudosimulation}. The resulting Extended PDM Score (EPDMS) is
\begin{equation}
\mathrm{EPDMS}
=\mathrm{NC}\cdot\mathrm{DAC}\cdot\mathrm{DDC}\cdot\mathrm{TLC}\cdot
\frac{5\,\mathrm{EP}+5\,\mathrm{TTC}+2\,\mathrm{LK}+2\,\mathrm{HC}+2\,\mathrm{EC}}{16}.
\label{eq:epdms}
\end{equation}
We report scores from the original evaluator as EPDMS$^{*}$ and those from the corrected official evaluator as EPDMS. All component scores in Equation~\eqref{eq:epdms} follow the official human-behavior filtering rule. A component is set to $1$ if its score for the human reference is $0$ and otherwise retains the model's score. Higher PDMS and EPDMS values indicate better planning performance.

\paragraph{HUGSIM evaluation.}
We evaluate zero-shot closed-loop generalization on all 436 HUGSIM scenarios without HUGSIM-specific training or fine-tuning~\citep{zhou2024hugsim}. HUGSIM defines No Collision (NC), Drivable Area Compliance (DAC), Time to Collision (TTC), Comfort (COM), Route Completion (RC), and the HUGSIM Driving Score (HD-Score). HD-Score aggregates the safety and comfort metrics over a closed-loop rollout and scales them by route completion.

We follow the HUGSIM evaluation protocol of WA-JEPA and take the baselines in Tab.~\ref{tab:hugsim-436} from its unified re-evaluation~\citep{wang2026wajepa}. We also report PDMS from the same evaluation protocol. The six metrics under All 436 Scenarios are averaged uniformly across scenarios. For each difficulty level, HD-Score is averaged within each source dataset and then uniformly across datasets. The overall HD-Score weights the Easy ($80$), Medium ($157$), Hard ($96$), and Extreme ($103$) difficulty levels according to their scenario counts.

\paragraph{Training configuration.}
We use AdamW for all three stages. Stage~1 distills DeiT-S from DINOv3 ViT-L/16, SigLIP2-SO400M NaFlex, and Depth Anything~3-Large (DA3) on NAVSIM front-view images. We weight the cosine and Smooth L1 feature losses by $0.9$ and $0.1$. Stage~2 initializes the encoder from Stage~1, while the register tokens, temporal attention, and planning modules are initialized from scratch. Stages~2 and 3 sample input frames at 2 Hz, corresponding to a temporal interval of 0.5 s. The backbone stays frozen and is adapted with rank-32 LoRA updates to the query and value projections. The trajectory and score losses have equal weights. The NC, DAC, DDC, TTC, EP, and comfort ranking weights are $1/1/0/5/5/2$. Stage~3 starts from the complete Stage~2 checkpoint and enables register-guided token compression. It uses the same optimizer settings as Stage~2 with reinitialized optimizer state. 

\paragraph{Efficiency profiling.}
We profile all methods with a batch size of one and four camera views on a single NVIDIA A100 GPU. We perform $100$ warm-up iterations followed by $1{,}000$ timed iterations under PyTorch inference mode. Each forward pass is measured with CUDA events and synchronized before its latency is recorded. The timed region covers model inference and excludes data loading and host-side preprocessing. We report mean end-to-end latency and compute throughput as $1{,}000$ divided by mean latency in milliseconds. Perception denotes the image encoder, planning denotes the trajectory generation and scoring modules, and remaining framework overhead is grouped under Others. Encoder FLOPs cover all camera views and frames, include both attention matrix multiplications, and use $1\,\mathrm{MAC}=2\,\mathrm{FLOPs}$.

\begin{table}[htbp]
\caption{\textbf{Hyperparameters for ConDenseDrive.}}
\label{tab:training-configuration}
\centering
\begingroup
\setlength{\tabcolsep}{14pt}
\begin{tabular}{@{}c l l@{}}
\toprule
\textbf{Stage} & \textbf{Parameter} & \textbf{Value} \\
\midrule
\multirow{6}{*}{1}
& Number of epochs & 50 \\
& Batch size & 128 \\
& Learning rate & $5\times10^{-4}$ \\
& Weight decay & $10^{-2}$ \\
& Warmup ratio & 0.10 \\
& Learning rate schedule & Cosine \\
\midrule
\multirow{6}{*}{2}
& Number of epochs & 25 \\
& Batch size & 64 \\
& Learning rate & $2\times10^{-4}$ \\
& Weight decay & $10^{-2}$ \\
& Warmup ratio & 0.10 \\
& Learning rate schedule & Cosine \\
\midrule
\multirow{8}{*}{3}
& Number of epochs & 10 \\
& Batch size & 64 \\
& Learning rate & $2\times10^{-4}$ \\
& Weight decay & $10^{-2}$ \\
& Warmup ratio & 0.10 \\
& Learning rate schedule & Cosine \\
& Compression blocks & 4, 7, 10 \\
\bottomrule
\end{tabular}
\endgroup
\end{table}

\FloatBarrier

\section{Additional Stage 1 Studies}
\label{app:stage1-transfer}
\paragraph{Encoder transferability.}
We further evaluate the distilled encoder with a camera-only DiffusionDrive planner. In Tab.~\ref{tab:diffusiondrive-encoder-transfer}, we replace its ResNet-34 backbone with our distilled DeiT-S while keeping all training settings and the remaining architecture unchanged. This improves PDMS from $86.0$ to $87.9$ without token compression. Both encoders have approximately $22$M parameters.

\begin{table}[!h]
\centering
\caption{\textbf{Encoder transfer to camera-only DiffusionDrive.}}
\label{tab:diffusiondrive-encoder-transfer}
\centering
\begingroup
\setlength{\tabcolsep}{3.0pt}
\resizebox{\linewidth}{!}{%
\begin{tabular}{l|l|ccccc|>{\columncolor{scoregraylight}}c}
\toprule
\textbf{Method}
& \textbf{Image Encoder}
& \textbf{NC}$\uparrow$
& \textbf{DAC}$\uparrow$
& \textbf{TTC}$\uparrow$
& \textbf{Comf.}$\uparrow$
& \textbf{EP}$\uparrow$
& \textbf{PDMS}$\uparrow$ \\
\midrule
DiffusionDrive ~\citep{liao2025diffusiondrive}
& ResNet-34 & \textbf{97.9} & 94.6 & 93.6 & \textbf{100} & 80.7 & 86.0 \\
DiffusionDrive + distilled encoder
& Distilled DeiT-S & \textbf{97.9} & \textbf{96.4} & \textbf{94.0} & \textbf{100} & \textbf{82.4} & \textbf{87.9} \\
\bottomrule
\end{tabular}%
}
\endgroup

\end{table}
\FloatBarrier

\section{Additional Stage 2 Studies}
\label{app:stage2-ablations}
\paragraph{Number of input frames.}
Tab.~\ref{tab:frame-ablation} compares dense models with different temporal lengths. Using two frames gives the best PDMS, while four frames increase computation without further improvement. We therefore use $T=2$ for the main model.

\begin{table}[!htbp]
\centering
\caption{\textbf{Effect of the number of input frames ($\rho=1.0$).}}
\label{tab:frame-ablation}
\centering
\begingroup
\setlength{\tabcolsep}{3.2pt}
\begin{tabular}{c|c|ccccc|>{\columncolor{scoregraylight}}c}
\toprule
\textbf{Frames}
& \textbf{FPS}$\uparrow$
& \textbf{NC}$\uparrow$
& \textbf{DAC}$\uparrow$
& \textbf{TTC}$\uparrow$
& \textbf{Comf.}$\uparrow$
& \textbf{EP}$\uparrow$
& \textbf{PDMS}$\uparrow$ \\
\midrule
$T=1$ & \textbf{25.4} & 99.1 & 98.8 & 96.6 & \textbf{100} & \textbf{91.2} & 94.1 \\
$T=2$ & 13.6 & \textbf{99.2} & \textbf{98.9} & \textbf{97.0} & \textbf{100} & \textbf{91.2} & \textbf{94.3} \\
$T=4$ & 7.3 & 99.1 & 98.5 & 96.8 & \textbf{100} & 91.1 & 93.9 \\
\bottomrule
\end{tabular}
\endgroup

\end{table}

\paragraph{View attention.}
The independently encoded register tokens also lack interaction across camera views. To study cross-view context, we arrange the view-wise register tokens at each time step:
\begin{equation}
    \mathbf{R}_{t}
    =\operatorname{ViewAttn}
    \!\left([\mathbf{R}_{t,1};\ldots;\mathbf{R}_{t,V}]\right).
    \label{eq:view_attention}
\end{equation}
This mirrors temporal attention in Equation~\eqref{eq:temporal_registers}, but groups camera views at a fixed time. Temporal attention reaches $94.3$ PDMS and performs best among the variants in Tab.~\ref{tab:attention-ablation}. Adding view attention does not improve the temporal model. Our final model therefore uses temporal attention without view attention and retains the registers from all input frames.

\begin{table}[!htbp]
\centering
\caption{\textbf{View and temporal attention variants ($\rho=1.0$).}}
\label{tab:attention-ablation}
\centering
\begingroup
\setlength{\tabcolsep}{3.0pt}
\begin{tabular}{l|c|ccccc|>{\columncolor{scoregraylight}}c}
\toprule
\textbf{Attention}
& \textbf{Frames}
& \textbf{NC}$\uparrow$
& \textbf{DAC}$\uparrow$
& \textbf{TTC}$\uparrow$
& \textbf{Comf.}$\uparrow$
& \textbf{EP}$\uparrow$
& \textbf{PDMS}$\uparrow$ \\
\midrule
None & $T=1$ & 99.1 & 98.8 & 96.6 & \textbf{100} & \textbf{91.2} & 94.1 \\
View & $T=1$ & 98.9 & 98.5 & 96.9 & \textbf{100} & 90.0 & 93.5 \\
\cmidrule(lr){1-8}
None & $T=2$ & 99.0 & 98.7 & 96.8 & \textbf{100} & 90.9 & 94.0 \\
Temporal & $T=2$ & \textbf{99.2} & \textbf{98.9} & \textbf{97.0} & \textbf{100} & \textbf{91.2} & \textbf{94.3} \\
View \& temporal & $T=2$ & 98.9 & 98.7 & 96.2 & \textbf{100} & 90.1 & 93.7 \\
\bottomrule
\end{tabular}
\endgroup

\end{table}

\paragraph{Register token scope.}
Tab.~\ref{tab:register-scope-ablation} compares the register tokens passed to the downstream planner. Using only the current-frame registers reaches $93.9$ PDMS, while retaining registers from all input frames reaches $94.3$. This result supports preserving the complete temporal register sequence after temporal attention.

\begin{table}[!htbp]
\centering
\begin{minipage}[t]{0.49\textwidth}
\vspace{0pt}
\centering
\caption{\textbf{Register scope ($T=2$, $\rho=1.0$).}}
\label{tab:register-scope-ablation}
\begingroup
\setlength{\tabcolsep}{3.0pt}
\renewcommand{\arraystretch}{1.08}
\resizebox{!}{1.70\baselineskip}{%
\begin{tabular}{l|ccccc|>{\columncolor{scoregraylight}}c}
\toprule
\textbf{Register Scope}
& \textbf{NC}$\uparrow$
& \textbf{DAC}$\uparrow$
& \textbf{TTC}$\uparrow$
& \textbf{Comf.}$\uparrow$
& \textbf{EP}$\uparrow$
& \textbf{PDMS}$\uparrow$ \\
\midrule
Current frame & 99.1 & 98.7 & 96.6 & \textbf{100} & 91.0 & 93.9 \\
All frames & \textbf{99.2} & \textbf{98.9} & \textbf{97.0} & \textbf{100} & \textbf{91.2} & \textbf{94.3} \\
\bottomrule
\end{tabular}%
}
\endgroup

\end{minipage}\hfill
\begin{minipage}[t]{0.49\textwidth}
\vspace{0pt}
\centering
\caption{\textbf{Token fusion ($T=1$, $\rho=0.7$).}}
\label{tab:token-fusion-ablation}
\centering
\begingroup
\setlength{\tabcolsep}{3.0pt}
\renewcommand{\arraystretch}{1.08}
\resizebox{!}{1.70\baselineskip}{%
\begin{tabular}{l|ccccc|>{\columncolor{scoregraylight}}c}
\toprule
\textbf{Token Handling}
& \textbf{NC}$\uparrow$
& \textbf{DAC}$\uparrow$
& \textbf{TTC}$\uparrow$
& \textbf{Comf.}$\uparrow$
& \textbf{EP}$\uparrow$
& \textbf{PDMS}$\uparrow$ \\
\midrule
Discard & \textbf{98.9} & 98.9 & 96.1 & \textbf{100} & 91.7 & 94.1 \\
Fuse & \textbf{98.9} & \textbf{99.0} & \textbf{96.3} & \textbf{100} & \textbf{91.8} & \textbf{94.3} \\
\bottomrule
\end{tabular}%
}
\endgroup

\end{minipage}
\end{table}

\section{Additional Stage 3 Studies}
\label{app:stage3-compression}
\paragraph{Token fusion.}
Tab.~\ref{tab:token-fusion-ablation} compares token fusion with direct token discarding. Fusion reaches $94.3$ PDMS, compared with $94.1$ for discarding. This result supports retaining a compact summary of low-scoring patch tokens during compression.

\paragraph{Keep rate.}
Tab.~\ref{tab:single-frame-keep-rate-ablation} reports the keep rate sweep in the single-frame setting. At $\rho=0.7$, encoder computation falls by about 41\% while PDMS improves from 94.1 to 94.3. At $\rho=0.3$, throughput reaches 44.5 FPS with only a 0.1 PDMS reduction from the dense model.

\begin{table}[!htbp]
\centering
\caption{\textbf{Keep rate ($T=1$).}}
\label{tab:single-frame-keep-rate-ablation}
\centering
\begingroup
\setlength{\tabcolsep}{5.0pt}
\renewcommand{\arraystretch}{1.08}
\begin{tabular}{c|cc|>{\columncolor{scoregraylight}}c}
\toprule
\textbf{Rate}
& \textbf{Encoder GFLOPs}$\downarrow$
& \textbf{FPS}$\uparrow$
& \textbf{PDMS}$\uparrow$ \\
\midrule
$\rho=1.0$ & 1181.77 & 25.4 & 94.1 \\
$\rho=0.9$ & 975.85 & 29.5 & 94.2 \\
$\rho=0.8$ & 818.59 & 32.6 & 94.2 \\
$\rho=0.7$ & 699.02 & 35.1 & \textbf{94.3} \\
$\rho=0.6$ & 607.73 & 37.4 & 94.1 \\
$\rho=0.5$ & 537.72 & 40.2 & 94.2 \\
$\rho=0.4$ & 484.30 & 42.6 & 94.1 \\
$\rho=0.3$ & \textbf{443.48} & \textbf{44.5} & 94.0 \\
\bottomrule
\end{tabular}
\endgroup

\end{table}
\FloatBarrier

\section{Additional Qualitative Visualizations}
\label{app:additional-qualitative-visualizations}

\begin{figure}[htbp]
    \centering
    \includegraphics[width=\linewidth]{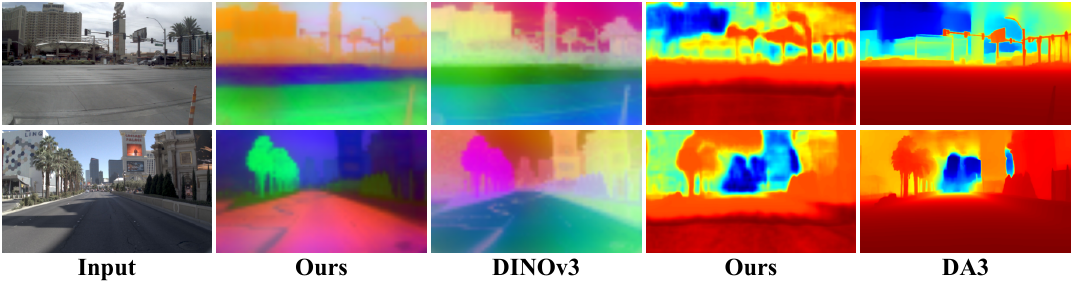}
    \caption{\textbf{PCA and depth visualizations.} 
    }
    \label{fig:additional-feature-visualizations}
\end{figure}

\paragraph{Feature Visualizations.}
Fig.~\ref{fig:additional-feature-visualizations} compares the student with DINOv3 and DA3 across two additional driving scenes.

\begin{figure}[htbp]
    \centering
    \includegraphics[width=\linewidth]{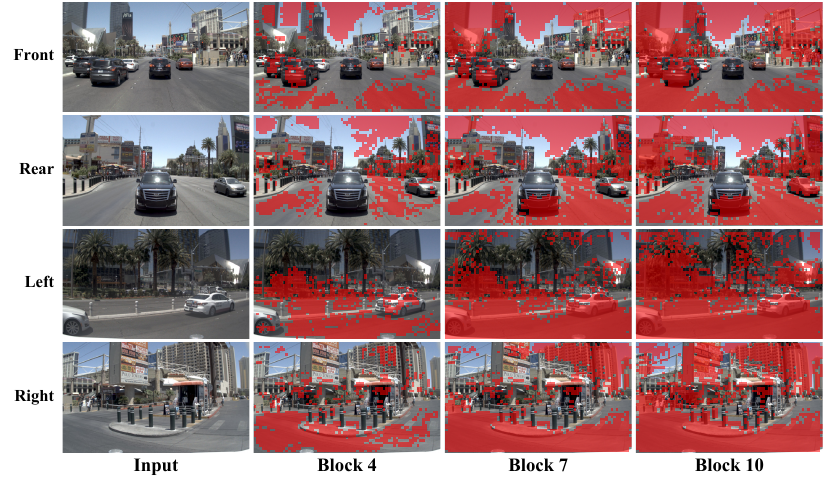}
    \caption{\textbf{Progressive register-guided token compression, Scene 1.} Results use $\rho=0.7$.}
    \label{fig:additional-token-compression-scene1}
\end{figure}

\begin{figure}[htbp]
    \centering
    \includegraphics[width=\linewidth]{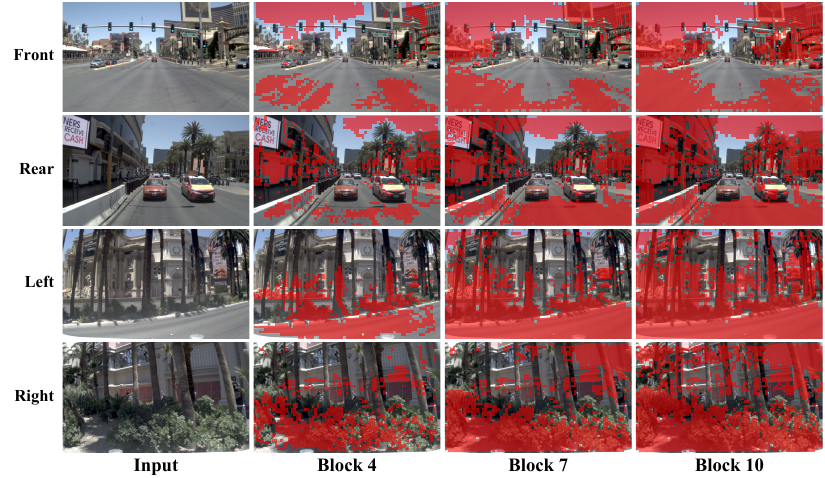}
    \caption{\textbf{Progressive register-guided token compression, Scene 2.} Results use $\rho=0.7$.}
    \label{fig:additional-token-compression-scene2}
\end{figure}

\paragraph{Token Compression Visualizations.}
Figs.~\ref{fig:additional-token-compression-scene1}--\ref{fig:additional-token-compression-scene2} show register-guided token compression across four camera views in two additional scenes.

\end{document}